\documentclass[10pt, epsf, rotating, pdftex, lscape, letterpaper]{article}
\usepackage[margin=1in]{geometry} 

\usepackage{lineno, amsmath}
\usepackage{booktabs} 
\usepackage{multirow}
\usepackage{soul} 
\usepackage{microtype}
\usepackage{amsmath}
\usepackage[ruled,lined,boxed,commentsnumbered]{algorithm2e}
\usepackage{cite}
\usepackage{amsmath,amssymb,amsfonts}
\usepackage[figuresright]{rotating}                                                         
\usepackage{soul}                                                             
\usepackage{microtype}                                                     
\usepackage{caption}                                                        

\begin{document}

\begin{center}
{\large \bf Semi-steady-state Jaya Algorithm}

\vspace{1cm}
{\bf U. K. Chakraborty}\\

Department of Computer Science\\
University of Missouri, St. Louis, MO 63121, USA\\
chakrabortyu@umsl.edu\\
\end{center}

\vspace{0.5cm}

\noindent 

\abstract{The Jaya algorithm is arguably one of the fastest-emerging metaheuristics amongst the newest members of the evolutionary computation family. The present paper proposes a new, improved Jaya algorithm by modifying the update strategies of the best and the worst members in the population.  Simulation results on a twelve-function benchmark test-suite as well as a real-world problem of practical importance show that the proposed strategy produces results that are better and faster in the majority of cases. Statistical tests of significance are used to validate the performance improvement.
}

\section{Introduction}
\label{sec:introduction}
For optimization of computationally hard problems and of problems that are mathematically intractable,  machine-learning-based strategies such as evolutionary computation (EC) \cite{michalewicz2013genetic} and artificial neural network (ANN) \cite{goodfellow2016deep} have seen significant success in numerous  application areas. 
The ``no-free-lunch theorem" \cite{wolpert1997no} tells us that, theoretically, over all possible optimization functions, all algorithms perform equally well. In practice, however, for specific problems (particularly, hard problems), the need for better and still better algorithms (and heuristics) remains.  
   
The Jaya algorithm \cite{rao2016jaya}, one of the newest members of the evolutionary computation family, has seen remarkable success across a wide variety of applications in continuous optimization (see Section~\ref{sec:survey} below). Jaya's success can arguably be attributed to the following two features: (a) it requires very few algorithm parameters, and (b) compared to most of its EC-cousins, Jaya is extremely simple to implement. A user of the Jaya algorithm has to decide on suitable values for only two parameters -- population size and the number of iterations (generations). Because any population-based algorithm (or heuristic) must have a population size, and because the user of any algorithm/heuristic must have an idea of when to stop the process, it can be argued that the population size and the stopping condition are two fundamental attributes of any population-based heuristic and that the Jaya algorithm is parameterless. 
In this paper, we present an algorithm that improves over the Jaya algorithm by modifying the search strategy, without compromising on the above two qualities. The comparative performance of Jaya and the proposed method is studied empirically on a twelve-function benchmark test-suite as well as on a real-world problem from fuel cell stack design optimization. The improvement in performance afforded by the proposed algorithm is validated with statistical tests of significance.
 (Technically, Jaya is not an algorithm; it is a heuristic. However, following common practice in the evolutionary computation community, we continue to refer to it as an algorithm in this paper.) 

The remainder of this paper is organized as follows. A very brief outline of some of the most interesting previous work on the Jaya algorithm is presented in Section~\ref{sec:survey}. Section~\ref{sec:proposed} presents the proposed algorithm. Simulation results and statistical tests for performance analysis are presented in Section~\ref{sec:simu}. Finally, conclusions are drawn in Section~\ref{sec:conclu}.

\section{A brief overview of previous work on Jaya} 
\label{sec:survey}

A variation of the standard Jaya algorithm is presented in the multi-team perturbation-guiding Jaya (MTPG-Jaya) \cite{rao2018multi} where several ``teams" explore the search space, with the same population being used by each team, while the ``perturbations" governing the progression of the teams are different. The MTPG-Jaya was applied to the layout optimization problem of a wind farm. 
The Jaya algorithm was originally designed for continuous (real-valued) optimization, and most of Jaya's applications to date have been in the continuous domain. A binary version of Jaya, however, was proposed   in \cite{li2017application}, where the authors borrowed (from \cite{pampara2005combining}) the idea of combining particle swarm optimization with angle modulation and adapted that idea for Jaya. The binary Jaya was applied to feature selection in \cite{li2017application}. 
Modifications to the standard Jaya algorithm include a self-adaptive multi-population-based Jaya algorithm 
that was applied to entropy generation minimization of a plate-fin heat exchanger \cite{rao2017self}, 
a multi-objective Jaya algorithm that was applied to waterjet machining process optimization \cite{rao2019multi}, and a hybrid parallel Jaya algorithm for a multi-core environment \cite{michailidis2019efficient}. Application areas of the Jaya algorithm have included such diverse fields as pathological brain detection systems \cite{nayak2018development}, flow-shop scheduling \cite{buddala2018improved}, maximum power point tracking problems in photovoltaic systems \cite{huang2017prediction}, identification and monitoring of electroencephalogram-based brain-computer interface for motor imagery tasks \cite{sinha2016jaya}, and traffic signal control \cite{gao2016jaya}.

\section{The proposed algorithm} 
\label{sec:proposed} 
The new algorithm is presented in Algorithm~\ref{algonew} where, without loss of generality, an array representation with conventional indexed access is assumed for the members (individuals) of a population. At each generation, we examine the individuals in the population one by one, in sequence, conditionally replacing each with a newly created individual. A new individual is created from the current individual by using the best individual, the worst individual, and two random numbers -- each chosen uniformly randomly in (0, 1] -- per problem parameter (variable). The generation of the new individual $x^{\mathrm{new}}$, given the current individual $x^{\mathrm{current}}$, is described by the following equation ($x^{\mathrm{new}}$, $x^{\mathrm{current}}$, $x^{\mathrm{best}}$ and $x^{\mathrm{new}}$ are each a $d$-component vector):
\begin{displaymath}
x_{i}^{\mathrm{new}} = x_{i}^{\mathrm{current}} + r_{t, i, 1} (x_{i}^{\mathrm{best}} -  |x_{i}^{\mathrm{current}}|) -  r_{t, i, 2} (x_{i}^{\mathrm{worst}} - |x_{i}^{\mathrm{current}}|)
\end{displaymath} 
where $x_{i}$, $i$ = 1 to $d$, represent the $d$ parameters (variables) to be optimized,  
$ r_{t, i, 1}$ and $ r_{t, i, 2}$ are each a random number in (0.0, 1.0], $t$ indicates the iteration (generation) number,
$x^{\mathrm{best}}$ and $x^{\mathrm{worst}}$ represent, respectively, the best and the worst individual in the population at the time of the creation of $x^{\mathrm{new}}$ from $x^{\mathrm{current}}$. When $x_{i}^{\mathrm{new}}$ falls outside its problem-specified lower or upper bound, it is clamped at the appropriate bound.

In the original Jaya algorithm, the new individual replaces the current individual only if it (the former) is better than the latter. The present algorithm, however, accepts the new individual if it is at least as good as the current individual.    

The original Jaya updates the population-best and the population-worst individuals once every generation. Algorithm~\ref{algonew}, however, checks to see if $x_{i}^{\mathrm{best}}$ needs to be updated, and performs the update if needed, after every single replacement of the existing individual.  
A similar approach is adopted for updating $x_{i}^{\mathrm{worst}}$, but in this case, an update is needed only for the case when the existing (current) individual is the worst one; this is because a replacement is guaranteed never to cause the objective (cost) function to be worse. 

The simultaneous presence in the population of more than one best (or worst) individual (clones of the same individual and/or different genotypes with the same phenotype) presents no problem for the new algorithm, because the computation of the best (or worst) is always over the entire population, that is, it is never done incrementally.  

We improve upon Jaya by changing the policies of updating the best and the worst members and also by changing the criterion used to accept a new member as a replacement of an existing member.
The motivation for the first pair of changes comes from the argument that an early availability and use of the best and worst individuals should lead to an earlier creation of better individuals; this is similar to the idea behind the ``steady-state" operation of genetic algorithms \cite{syswerda1991study, chakraborty1996analysis}.  The logic behind the second change is to try to avoid the ``plateau problem". 

We call the proposed algorithm semi-steady-state Jaya or SJaya.

\begin{algorithm}
initialize the population\;
find the best and the worst individuals in the population, and~initialize {\it bestIndex} to the index of the best individual and {\it worstIndex} to the index of the worst individual\;
\While{a pre-determined stopping condition is not satisfied}{
set the parameters (the $r$'s), independently of one another, to~random values between 0.0 and 1.0\;
\For{each individual in the population starting from the first index}{
 create a new individual using the current individual, the~individual at {\it bestIndex}, the~individual at {\it worstIndex}, and~the random parameters\;

\If{the new individual is at least as good as the current individual}{
replace the current individual with the new individual\;
\If{the current individual is better than the individual at bestIndex}{
update {\it bestIndex} to set it to the current index\;
}
\If{the current individual's index is the same as {\it worstIndex}}{
find the worst individual in the population and set {\it worstIndex} to the index of the worst individual\;  
}
}
}
}
\caption{Pseudocode of the improved algorithm.}
\label{algonew}
\end{algorithm}

\section{Simulation results} \label{sec:simu}
For studying the comparative performance of Jaya and SJaya, we use a benchmark test-suite comprising a dozen well-known test functions from the literature and a real-world problem of fuel cell stack design optimization. All of the thirteen problems involve minimization of the objective function value (fitness). The following metrics \cite{chakraborty2012pem} are used for performance comparison:
\begin{itemize}
\item Best-of-run fitness: the best (lowest), mean, and standard deviation (over 30 runs) of the best-of-run fitness values;
\item The number of fitness evaluations (FirstHitEvals) needed to reach a specified fitness value for the first time in a run:  the best (fewest), mean, and standard deviation (over 30 runs) of these numbers;
\item Success count: The number of runs (out of the thirty) in which the specified fitness level is reached (it is possible that the specified level is never reached with the given population size and the given number of generations).
\end{itemize}
The best-of-run fitness provides a measure of the quality of the solution, while the FirstHitEvals metric expresses how fast the algorithm is able to find a solution of a given quality. The two metrics are thus complementary to each other.  

\subsection{Results on the benchmark test-suite}
The benchmark suite (Table~\ref{tab:bench}) includes functions of a wide variety of features and levels of problem difficulty, including unimodal/multimodal, separable/non-separable, continuous/discontinuous, differentiable/non-differentiable, and convex/non-convex functions. 

\begin{sidewaystable*}
\captionsetup{justification=centering}
\caption{Benchmark functions.}
\label{tab:bench}
\centering
\begin{tabular}{lllll@{}} 
\toprule
\textbf{Name} & \textbf{Definition} & \textbf{Dim.} & \textbf{Global Minimum} 	& \textbf{Bounds}\\
\midrule
Ackley   & $f(x_1, \cdots, x_n) = -20 exp\left(-0.2 \sqrt{\frac{1}{n} \sum_{i=1}^n x_i^2}\right) -exp\left(\frac{1}{n} \sum_{i=1}^n cos(2\pi x_i)\right) + 20 + e$ & 
30 &
\begin{tabular}{l}
  $f(x^{*}) = 0$ \\
  $x^{*} = (0, \cdots, 0)$ \\
\end{tabular} 
& $-10 \le x_{i} \le 10$ \\[4mm]

Rosenbrock  & 
$f(x_1, \cdots, x_n) = \sum_{i=1}^{n-1}[100 (x_{i+1} - x_i^2)^ 2 + (1 - x_i)^2]$ 
& 30 &
\begin{tabular}{l}
  $f(x^{*}) = 0$ \\
  $x^{*} = (1, \cdots, 1)$ \\
\end{tabular} 
& $-10 \le x_{i} \le 10$ \\[4mm]

Chung-Reynolds   & 
$f(x_1, \cdots, x_n) = \left(\sum_{i=1}^{n} x_{i}^{2}\right)^{2}$ 
& 30 &
\begin{tabular}{l}
  $f(x^{*}) = 0$ \\
  $x^{*} = (0, \cdots, 0)$ \\
\end{tabular} 
& $-10 \le x_{i} \le 10$ \\[4mm]

Step   & 
$f(x_1, \cdots, x_n) = \sum_{i=1}^{n} \lfloor|x_{i}|\rfloor$ 
& 30 &
\begin{tabular}{l}
  $f(x^{*}) = 0$ \\
  $x_{i}^{*} \in (-1, 1)$ \\
\end{tabular} 
& $-100 \le x_{i} \le 100$ \\[4mm]

Alpine-1   & 
$f(x_1, \cdots, x_n) = \sum_{i=1}^{n} |x_{i}\sin(x_{i}) + 0.1x_{i}|$ 
& 30 &
\begin{tabular}{l}
  $f(x^{*}) = 0$ \\
  $x^{*} = (0, \cdots, 0)$ \\
\end{tabular} 
& $-10 \le x_{i} \le 10$ \\[4mm]

SumSquares   & 
$f(x_1, \cdots, x_n) = \sum_{i=1}^{n} ix_{i}^{2}$ 
& 30 &
\begin{tabular}{l}
  $f(x^{*}) = 0$ \\
  $x^{*} = (0, \cdots, 0)$ \\
\end{tabular} 
& $-10 \le x_{i} \le 10$ \\[4mm]

Sphere   & 
$f(x_1, \cdots, x_n) = \sum_{i=1}^{n} x_{i}^{2}$ 
& 30 &
\begin{tabular}{l}
  $f(x^{*}) = 0$ \\
  $x^{*} = (0, \cdots, 0)$ \\
\end{tabular} 
& $-100 \le x_{i} \le 100$ \\[4mm]

Bohachevsky-3  & 
$f(x_1, x_2) =  x_{1}^{2} + 2x_{2}^{2} - 0.3\cos(3 \pi x_{1}+ 4 \pi x_{2}) +0.3$ 
& 2 &
\begin{tabular}{l}
  $f(x^{*}) = 0$ \\
  $x^{*} = (0, 0)$ \\
\end{tabular} 
& $-100 \le x_{1}, x_{2} \le 100$ \\[4mm]

Bohachevsky-2   & 
$f(x_1, x_2) =  x_{1}^{2} + 2x_{2}^{2} - 0.3\cos(3 \pi x_{1}) \cos(4 \pi x_{2}) +0.3$ 
& 2 &
\begin{tabular}{l}
  $f(x^{*}) = 0$ \\
  $x^{*} = (0, 0)$ \\
\end{tabular} 
& $-100 \le x_{1}, x_{2} \le 100$ \\[4mm]

Bartels Conn   & 
$f(x_1, x_2) =  |x_{1}^{2} + x_{2}^{2} + x_{1}x_{2}| + |\sin(x_{1})| + |\cos(x_{2})|$ 
& 2 &
\begin{tabular}{l}
  $f(x^{*}) = 1$ \\
  $x^{*} = (0, 0)$ \\
\end{tabular} 
& $-500 \le x_{1}, x_{2} \le 500$ \\[4mm]

Goldstein-Price  & 
\begin{tabular}{lll}
$f(x_1, x_2)$ & = & $\left[1 + (x_1 + x_2 + 1)^2(19 - 14x_1 +3x_1^2 - 14x_2 + 6x_1 x_2 + 3x_2^2)\right] \times$  \\
                &    & $\left[30 + (2x_1 - 3x_2)^2 (18 - 32x_1 + 12x_1^2 + 48x_2 - 36x_1 x_2 + 27x_2^2)\right]$ 
\end{tabular}
& 2 &
\begin{tabular}{l}
  $f(x^{*}) = 3$ \\
  $x^{*} = (0, -1)$ \\
\end{tabular} 
& $-2 \le x_{1}, x_{2} \le 2$ \\[4mm]

Matyas   & 
$f(x_1, x_2) =  0.26 (x_{1}^{2} + x_{2}^{2}) - 0.48 x_{1} x_{2}$ 
& 2 &
\begin{tabular}{l}
  $f(x^{*}) = 0$ \\
  $x^{*} = (0, 0)$ \\
\end{tabular} 
& $-10 \le x_{1}, x_{2} \le 10$ \\[4mm]

\bottomrule
\end{tabular}
\end{sidewaystable*}

For each test function, the population size and the number of generations were chosen based loosely on the problem size (number of variables) and the problem difficulty. No systematic tuning of the population size (PopSize) or the number of generations  (Gens) was attempted; the values used in this study were found to be reasonably good across a majority of the problems after a few initial trials. Two PopSize-Gens combinations were used for each function (see Table~\ref{tab:sssjaya}). For $d$ = 30, population sizes of 100 and 150 were used, with the corresponding number of generations being 3000 and 5000. For $d$ = 2, the population sizes were 15 and 20, with 5000 generations used for both.
Thirty independent runs of each of the two algorithms were executed for each PopSize-Gens combination on each of the test functions. A run is considered a success if it manages to produce at least one solution with a fitness within a distance of $\pm$1.0e-6 from the true (known) global optimum, and the number of fitness evaluations corresponding to the first appearance of such a solution is recorded as the FirstHitEvals of that run.   

Tables~\ref{tab:sssjaya}~and~\ref{tab:jaya} show the results of SJaya and Jaya, respectively, on the 12-function test-suite. In all the tables in this paper results are rounded at the fourth decimal place.

\begin{sidewaystable*}
\caption{Results of SJaya on the 12-function test-suite (each row corresponds to 30 independent runs). Most numbers are shown with rounding at the fourth place after the decimal.}
\label{tab:sssjaya}
\centering
\begin{tabular}{@{\extracolsep{5pt}}cccccccccc@{}} 
\toprule
 \multirow{2}{*}{\textbf{Function}} & \multirow{2}{*}{\textbf{PopSize}} & \multirow{2}{*}{\textbf{Gens}} 	& \multicolumn{3}{c} {\bf Best-of-run Fitness} & \multicolumn{4}{c} {\bf FirstHitEvals}\\
\cmidrule{4-6}  \cmidrule{7-10}                                               
                          &                         &                        &  \textbf{Best} & \textbf{Mean} & \textbf{Std Dev} &  \textbf{Success} &\textbf{Best} & \textbf{Mean} & \textbf{Std Dev} \\
\midrule
\multirow{2}{*}{Ackley} & 100 & 3000   &   7.4347e-10 & 1.8090e-09 & 9.1920e-10 & 30 & 209499 & 217209.4333 & 4885.3830 \\
                               & 150 & 5000   &  1.0938e-12 & 2.7097e-12 & 7.9283e-13  & 30 & 407146 & 426516.7667 & 7522.8400 \\[2mm]
\multirow{2}{*}{Rosenb} & 100 & 3000   & 0.0015 & 25.4532 & 28.8764              &  0  &    ---          &   ---                &   ---              \\
                                   & 150 & 5000   & 0.0001 & 17.0565 & 26.9145              & 0   &    ---          &   ---                &   ---   \\[2mm]

\multirow{2}{*}{Chu-Rey} & 100 & 3000      &  5.0261e-37 & 1.1798e-35 & 3.0313e-35 & 30 & 77035 & 84420.6 & 3325.8495  \\ 
                                      & 150 & 5000      &   1.2691e-48 & 4.9288e-47 & 6.4529e-47 & 30 & 153594 & 162497.0667  & 3651.8492  \\[2mm]

\multirow{2}{*}{Step} & 100 & 3000      &  0.0 & 0.0667 & 0.2494  & 28 & 39004 & 43895.0357  & 5319.6538  \\ 
                                 & 150 & 5000    &  0.0 & 0.0 & 0.0 &  30 & 68099 & 73154.9333 & 3639.5311  \\ [2mm]

\multirow{2}{*}{Alp-1} & 100 & 3000     &   0.0247  & 6.8245  & 6.4345  &  0  &    ---          &   ---                &   ---  \\ 
                               & 150 & 5000     & 0.0137  & 4.5976  & 5.7499 &  0  &    ---          &   ---                &   --- \\[2mm]

\multirow{2}{*}{F2-Rao} & 100 & 3000     &  6.1724e-18 & 3.8440e-17 & 4.0234e-17 & 30 & 138646 & 144029.4333  & 3771.9204  \\  
                                    & 150 & 5000     &  1.3309e-23 & 7.2599e-23 & 5.5164e-23 &  30 & 266653 & 280539.4333  & 6344.5950  \\[2mm] 

\multirow{2}{*}{Sphere} & 100 & 3000     &  5.6616e-17 & 2.9297e-16 & 2.6115e-16 & 30 & 152133 & 157149.2333  & 2954.1983   \\ 
                                    & 150 & 5000     &  1.3981e-22 & 6.1597e-22 & 4.1632e-22 & 30 & 298554 & 306880.0667  & 4927.0814  \\ [2mm]   

\multirow{2}{*}{Boha-3}  & 15 & 5000      &  0.0 & 0.0 & 0.0 & 30 & 882 & 1322.4667  & 308.4498   \\
                                     & 20 & 5000      &  0.0 & 0.0 & 0.0 &  30 & 1182 & 1838.7 & 333.6645  \\[2mm]

\multirow{2}{*}{Boha-2} & 15 & 5000      &   0.0 & 0.0 & 0.0 &  30 & 718 & 1005.3333  & 268.2153  \\
                                    & 20 & 5000      &  0.0 & 0.0 & 0.0 & 30 & 890 & 1443.3667  & 222.3957  \\ [2mm]

\multirow{2}{*}{Bartel} & 15 & 5000      & 1.0 & 1.0 & 0.0 & 30 & 893 & 1061.0 & 90.1706   \\
                                  & 20 & 5000       & 1.0 & 1.0 & 0.0 & 30 & 1128 & 1523.4333  & 124.4451   \\[2mm]
                          
\multirow{2}{*}{Gold-P} & 15 & 5000      &  3.0000  & 3.0000 & 1.0820e-05 &  6 & 28320 & 55587.5 & 14917.3860   \\
                                   & 20 & 5000      &  3.0000  & 3.0000  & 1.8986e-05 & 5 & 58442 & 82977.0 & 14243.2530   \\[2mm]
  
\multirow{2}{*}{Matyas} & 15 & 5000      &  0.0 & 3.0482e-35 & 1.6415e-34 &  30 & 471 & 856.1 & 169.1497  \\
                                    & 20 & 5000      &   0.0 & 5.6005e-123 & 3.0160e-122 & 30 & 692 & 1152.7333  & 264.9280   \\[2mm]

\bottomrule
\end{tabular}
\end{sidewaystable*}

\begin{sidewaystable*}
\caption{Results of Jaya on the 12-function test-suite (each row corresponds to 30 independent runs). Most numbers are shown with rounding at the fourth place after the decimal.}
\label{tab:jaya}
\centering
\begin{tabular}{@{\extracolsep{5pt}}cccccccccc@{}} 
\toprule
 \multirow{2}{*}{\textbf{Function}} & \multirow{2}{*}{\textbf{PopSize}} & \multirow{2}{*}{\textbf{Gens}} 	& \multicolumn{3}{c} {\bf Best-of-run Fitness} & \multicolumn{4}{c} {\bf FirstHitEvals}\\
\cmidrule{4-6}  \cmidrule{7-10}                                               
                          &                         &                        &  \textbf{Best} & \textbf{Mean} & \textbf{Std Dev} &  \textbf{Success} &\textbf{Best} & \textbf{Mean} & \textbf{Std Dev} \\
\midrule
\multirow{2}{*}{Ackley} & 100 & 3000   &  4.2232e-06 & 7.6506e-06 & 1.9595e-06  &  0  &    ---          &   ---                &   ---        \\
                                   & 150 & 5000   &  3.9148e-08 & 8.2624e-08 & 2.5913e-08 & 30 & 620422 & 651813.4333  & 11801.5819  \\[2mm]
\multirow{2}{*}{Rosenb} & 100 & 3000  &  0.0310  & 26.8113  & 27.5200  &   0  &    ---          &   ---                &   ---       \\
                                   & 150 & 5000   &  0.0521  & 37.0939  & 32.6063  & 0  &    ---          &   ---                &   --- \\[2mm]

\multirow{2}{*}{Chu-Rey} & 100 & 3000      &    6.1251e-23 & 2.2695e-21 & 2.7432e-21 & 30 & 122216 & 130083.4667  & 3283.9261  \\ 
                                      & 150 & 5000      &  1.1798e-30 & 1.1626e-29 & 1.2429e-29 & 30 & 230733 & 245191.6 & 6139.8179     \\[2mm]

\multirow{2}{*}{Step} & 100 & 3000      &   0.0 & 0.0 & 0.0 & 30 & 82115 & 88940.6 & 4467.5422   \\ 
                                 & 150 & 5000    &   0.0 & 0.0 & 0.0 & 30 & 154374 & 166652.7667  & 6105.3915  \\ [2mm]

\multirow{2}{*}{Alp-1} & 100 & 3000     &    0.0240  & 9.7502  & 5.6913  & 0  &    ---          &   ---                &   ---    \\ 
                               & 150 & 5000     &   0.0381  & 6.2610  & 5.6690  & 0  &    ---          &   ---                &   --- \\[2mm]

\multirow{2}{*}{F2-Rao} & 100 & 3000     &    1.5297e-10 & 4.5700e-10 & 2.2292e-10 & 30 & 213427 & 222775.1667  & 3910.2976   \\  
                                    & 150 & 5000     & 4.9038e-15 & 3.7103e-14 & 1.7512e-14 &  30 & 406918 & 421195.1 & 7055.2581  \\[2mm] 

\multirow{2}{*}{Sphere} & 100 & 3000     &    1.2410e-09 & 4.6650e-09 & 2.4779e-09 & 30 & 231137 & 245599.1667  & 4874.0277  \\ 
                                    & 150 & 5000     &  8.6939e-14 & 3.6152e-13 & 2.3875e-13 & 30 & 441574 & 464684.3667  & 10701.7923     \\[2mm]    

\multirow{2}{*}{Boha-3}  & 15 & 5000      &  0.0 & 0.0301  & 0.1624  &  29 & 947 & 1368.5517  & 257.8614   \\
                                     & 20 & 5000      &  0.0 & 0.0 & 0.0 & 30 & 1461 & 1877.5333  & 275.5259    \\[2mm]

\multirow{2}{*}{Boha-2} & 15 & 5000      &   0.0 & 0.0347  & 0.1866  &  29 & 809 & 1102.7931  & 158.0520    \\
                                    & 20 & 5000      &   0.0 & 0.0 & 0.0 &  30 & 1160 & 1590.8667  & 243.5768   \\ [2mm]

\multirow{2}{*}{Bartel} & 15 & 5000      &   1.0 & 1.0 & 0.0 &  30 & 995 & 1238.7667  & 91.8632  \\
                                  & 20 & 5000       &   1.0 & 1.0 & 0.0 &  30 & 1375 & 1684.0667  & 152.4998  \\[2mm]
                          
\multirow{2}{*}{Gold-P} & 15 & 5000      &   3.0000  & 3.0000  & 1.4203e-05 &  5 & 36981 & 57683.4 & 12921.8507   \\
                                   & 20 & 5000      &    3.0000  & 3.0000  & 1.7344e-05 & 3 & 37550 & 52030.0 & 15017.5414  \\[2mm]
  
\multirow{2}{*}{Matyas} & 15 & 5000      &   0.0 & 1.6173e-11 & 8.7092e-11 & 30 & 572 & 906.9667  & 261.1821   \\
                                    & 20 & 5000      &   0.0 & 1.9566e-55 & 1.0537e-54 & 30 & 761 & 1286.0 & 264.6156     \\[2mm]

\bottomrule
\end{tabular}
\end{sidewaystable*}

From Tables~\ref{tab:sssjaya}~and~\ref{tab:jaya} we see that SJaya produces superior results than Jaya on all the metrics. Specifically,
\begin{itemize}
\item On the best of best-of-runs metric, out of 24 cases, SJaya outperforms Jaya in 12 cases and is outperformed by Jaya in 2 cases, with 10 cases resulting in ties. In a few cases (such as the values of 3.0000 of the best of best-of-run fitnesses and of the mean of best-of-run fitnesses corresponding to the Goldstein-Price function for both SJaya and Jaya), differences exist at the fifth or a later decimal position but do not show in Tables~\ref{tab:sssjaya}~and~\ref{tab:jaya}.
\item On the mean of best-of-runs metric, SJaya is the winner with win-loss-tie figures of 18-1-5. 
\item The success counts are higher (5-1-18) for SJaya. 
\item SJaya outperforms Jaya 19-1-4 on the best FirstHitEvals metric.
\item On the mean FirstHitEvals metric, SJaya outperforms Jaya 19-1-4.
\end{itemize}

Table~\ref{tab:tscores} presents the $t$-scores and one-tailed $p$-values from Smith-Satterthwaite tests (Welch's tests) \cite{johnson2000probability} (corresponding to unequal population variances) run on the data in Tables~\ref{tab:sssjaya}~and~\ref{tab:jaya} for examining whether or not the difference between the means of Jaya and SJaya (for the best-of-run fitnesses metric and, separately, for the FirstHitEvals metric) is significant. 
Using the subscripts 1 and 2 for Jaya and SJaya respectively, we obtain the 
test statistic as a $t$-score given by
\begin{displaymath}
t = \frac{\displaystyle\bar{x}_{1} - \bar{x}_{2} - 0}{\displaystyle\sqrt{\frac{s_{1}^{2}}{n_1} + \frac{s_{2}^{2}}{n_2}}},
\end{displaymath}
and the degrees of freedom of the $t$-distribution (this $t$-distribution is used to approximate the sampling distribution of the difference between the two means) as 
\begin{displaymath}
\frac{\displaystyle \left(\frac{s_1^2}{n_1}+\frac{s_2^2}{n_2}\right)^2}{\displaystyle \frac{(s_1^2/n_1)^2}{n_1-1}+\frac{(s_2^2/n_2)^2}{n_2-1}},
\end{displaymath}
where  the symbols $\bar{x}$, $s$ and $n$ represent mean, standard deviation and sample size, respectively. Note that even though 30 runs were executed in each case, the sample sizes are not always 30 (because not all runs were successful in all cases); for instance, for the Goldstein-Price function (executed with parameters PopSize = 15 and Gens = 5000),  $n_{1}$ = $n_{2}$ = 30 for the mean best-of-run fitness calculation, whereas   $n_{1}$ = 5 and $n_{2}$ = 6 for the mean FirstHitEvals computation.
(To avoid division by zero, we cannot use the above formulas when both $s_{1}$ and $s_{2}$ are zeros or when any one of $n_{1}$ and $n_{2}$ is unity.)

Using $\alpha$ = 0.05 as the level of significance, we see from the results in Table~\ref{tab:tscores} that on the best-of-run metric, out of a total of 19 cases, ten cases produce a positive $t$ statistic that corresponds to a one-tailed $p$-value less than $\alpha$ (the $p$-values were obtained with $t$-tests from {\em scipy.stats}). Thus the null hypothesis  $\bar{x}_{1} = \bar{x}_{2}$ must be rejected in favor of $\bar{x}_{1} > \bar{x}_{2}$ for those ten cases. The 19 cases include a lone negative $t$ score, but the corresponding $p$-value is greater than 0.05. On the FirstHitEvals metric, we have a total of 19 cases (the two occurrences of 19 between best-of-run and FirstHitEvals is a coincidence), of which fourteen have a positive $t$ with a $p$-value less than 0.05, and a single case has a negative $t$-score with a less-than-0.05 $p$-value.

\begin{table*}
\caption{Smith-Satterthwaite tests: Jaya vs. SJaya on the benchmark functions}
\label{tab:tscores}
\centering
\begin{tabular}{@{\extracolsep{5pt}}ccccccc@{}} 
\toprule
 \multirow{2}{*}{\textbf{Function}} & \multirow{2}{*}{\textbf{PopSize}} & \multirow{2}{*}{\textbf{Gens}} 	& \multicolumn{2}{c} {\bf Best-of-run Fitness} & \multicolumn{2}{c} {\bf FirstHitEvals}\\
\cmidrule{4-5}  \cmidrule{6-7}                                               
                          &                         &                        &  \textbf{$t$-statistic} &  \textbf{$p$-value} & \textbf{$t$-statistic} & \textbf{$p$-value} \\
\midrule
\multirow{2}{*}{Ackley} & 100 & 3000   & 21.3800 &  1.3355e-19  & --- & ---        \\
                                   & 150 & 5000   &   17.4636 & 3.1280e-17  & 88.1720  & 3.7508e-56  \\[2mm]
\multirow{2}{*}{Rosenb} & 100 & 3000  &    0.1865 & 0.4264  &   --- & --- \\
                                   & 150 & 5000   &  2.5958 & 0.0060   &   --- & --- \\[2mm]

\multirow{2}{*}{Chu-Rey} & 100 & 3000      & 4.5314 &  4.6542e-05  & 53.5110  & 2.3156e-51  \\ 
                                      & 150 & 5000      &  5.1236  &  8.9954e-06   & 63.4031  &  1.1548e-47    \\[2mm]

\multirow{2}{*}{Step} & 100 & 3000      &  -1.4639  & 0.0770  & 34.7952  & 1.9600e-38    \\ 
                                 & 150 & 5000    &    --- & --- &  72.0480  &   2.6003e-50  \\ [2mm]

\multirow{2}{*}{Alp-1} & 100 & 3000     &    1.8655  & 0.0336  &   --- & ---   \\ 
                                 & 150 & 5000     &   1.1283  &  0.1319  &   --- & --- \\[2mm]

\multirow{2}{*}{F2-Rao} & 100 & 3000     &  11.2285 & 2.2360e-12  & 79.3863  & 4.1571e-61    \\  
                                    & 150 & 5000     &  11.6045   &  1.0180e-12  & 81.1938  &  3.3244e-61   \\[2mm] 

\multirow{2}{*}{Sphere} & 100 & 3000     &    10.3116 & 1.6374e-11 &  85.0016  & 4.2333e-54  \\ 
                                    & 150 & 5000     &     8.2938  &   1.9158e-09   &   73.3631  &  3.1842e-45   \\[2mm]    

\multirow{2}{*}{Boha-3}  & 15 & 5000      &   1.0171 &  0.1588  & 0.6234 & 0.2678   \\
                                     & 20 & 5000      &    --- & --- & 0.4915  &  0.3125   \\[2mm]

\multirow{2}{*}{Boha-2} & 15 & 5000      &   1.0171  &  0.1588 &  1.7071  &  0.0472    \\
                                    & 20 & 5000      &    --- & --- & 2.4494  &  0.0087  \\ [2mm]

\multirow{2}{*}{Bartel} & 15 & 5000      &    --- & --- & 7.5641  &  1.6549e-10   \\
                                  & 20 & 5000       &   --- & --- &  4.4699  &  1.9412e-05   \\[2mm]
                          
\multirow{2}{*}{Gold-P} & 15 & 5000      &   1.0676  &  0.1452  & 0.2496  &  0.4042    \\
                                   & 20 & 5000      &   0.7407  &   0.2309  &  -2.8765  &  0.0217   \\[2mm]
  
\multirow{2}{*}{Matyas} & 15 & 5000      &  1.0171 & 0.1588  &  0.8954  &  0.1875    \\
                                    & 20 & 5000      &  1.0171 & 0.1588  &  1.9494  &  0.0280      \\[2mm]

\bottomrule
\end{tabular}
\end{table*}

The statistical tests in Table~\ref{tab:tscores} provide performance comparison separately on each of the twelve functions (using two different algorithm parameter settings for each function). A measure of the combined performance on the 12 functions taken together can be obtained using a paired-sample Wilcoxon signed rank test on the 12-function suite. The results of this test for each of the two metrics are presented in Table~\ref{tab:wilcoxon} where the null hypothesis is that the Jaya mean and the SJaya mean are identical and the alternate hypothesis is that the former is larger than the latter.
The second column in Table~\ref{tab:wilcoxon} shows the number of zero differences between SJaya and Jaya; $n$ represents the effective number of samples obtained by ignoring the samples, if any, corresponding to zero differences (e.g., $n$ is $24 - 5 = 19$ for the mean of best-of-run fitness metric); $W$ is the test statistic obtained as the minimum of $W+$ and $W-$; $\alpha$ represents the level of significance (a value of 0.05 is used here); and the critical $W$ for a given $n$ and for $\alpha$ = 0.05 is obtained from standard statistical tables. The $W$ statistic is seen to be less than the critical $W$. Arguing that the sample size is large enough for the discrete distribution of the $W$ statistic to be approximated by a continuous distribution, 
we obtain the mean of $W$ as   
\begin{displaymath}
\mathrm{mean} = \frac{n(n+1)}{4}, 
\end{displaymath}
and its standard deviation as
\begin{displaymath}
\mathrm{std\:dev}= \sqrt{\frac{n(n+1)(2n+1)}{24}}, 
\end{displaymath}
and, under the normal distribution assumption, the $z$-statistic is obtained from 
\begin{displaymath}
z = \frac{W - \mathrm{mean}}{\mathrm{std\:dev}}. 
\end{displaymath}
The one-tailed $p$-value corresponding to the above $z$-statistic is obtained from standard tables of the normal distribution.

From the results in Tables~\ref{tab:tscores} and \ref{tab:wilcoxon} we conclude that  at the 5\% significance level, SJaya is better than Jaya on the benchmark test-set.

\begin{sidewaystable}
\caption{Wilcoxon signed rank tests: Jaya vs. SJaya on the 12-function benchmark suite.}
\centering
\begin{tabular}{p{4.5cm} p{1cm} p{1cm} p{1cm} p{1cm} p{1cm} p{1cm} p{1cm} p{1cm} p{1.5cm} p{1.5cm} p{1cm}}
\toprule
\textbf{Metric} & \textbf{\#zero diff.} & \textbf{$n$} & \textbf{$W+$} &  \textbf{$W-$} &  \textbf{$W$} & \textbf{$\alpha$} & \textbf{Critical $W$} & \textbf{Mean of $W$} & \textbf{Std. Dev. of $W$} & \textbf{$z$-statistic} & \textbf{left tail $p$} \\ 
\midrule
Mean of Best-of-Run Fitnesses  & 5 & 19 & 175 & 15 & 15 & 0.05 & 53 & 95 & 24.8495 & $-$3.2194 & 0.0006 \\
Mean of FirstHitEvals               & 0 & 19 & 180 & 10 & 10 & 0.05 & 53 & 95 & 24.8495 & $-$3.4206  & 0.0003 \\
\bottomrule
\end{tabular}
\label{tab:wilcoxon}
\end{sidewaystable}

\subsection{Results on fuel cell stack design optimization}
A proton exchange membrane fuel cell (PEMFC) \cite{larminie2003fuel, o2016fuel} stack design optimization problem \cite{mohamed2004proton, chakraborty2019proton, besseris2014using} is considered here. This problem has been investigated in the fuel cell literature as a problem of practical importance for which the global minimum is believed to be mathematically intractable \cite{chakraborty2019proton}. 
This is a constrained optimization problem where the task is to minimize the cost of building a PEMFC stack that meets specific requirements. The objective (cost) function is a function of three variables $N_{\mathrm{p}}, N_{\mathrm{s}}, A_{\mathrm{cell}}$:
\begin{displaymath}
\mathrm{cost}  = 
K_{\mathrm{n}} \times N_{\mathrm{p}} \times N_{\mathrm{s}} + 
K_{\mathrm{diff}} \times 
\left|V_{\mathrm{load, rated}} - V_{\mathrm{load, mpp}} \right| + 
K_{\mathrm{a}} \times A_{\mathrm{cell}} + {\cal P},
\label{eq:cost}
\end{displaymath}
where 
$N_{\mathrm{s}}$ is the number of cells connected in series in each group; $N_{\mathrm{p}}$ is the number of groups connected in parallel; $A_{\mathrm{cell}}$ is the cell area;
$V_{\mathrm{load, r}}$  is the rated (given) terminal voltage of the stack;                   
$V_{\mathrm{load, mpp}}$  represents the output voltage at the maximum power point of the stack;                     
$P_{\mathrm{load, r}}$ is the rated (given) output power of the stack;                     
$P_{\mathrm{load, max}}$ is  the maximum output power of the stack; 
$K_{\mathrm{n}}, K_{\mathrm{diff}}, K_{\mathrm{a}}$ are pre-determined constants \cite{chakraborty2019proton} used to adjust the relative importance of the different components of the cost function;
and  ${\cal P}$ represents a penalty term given by
\begin{displaymath}
{\cal P} = \begin{cases}
                     0                             & \text{if $P_{\mathrm{load, max}} \ge P_{\mathrm{load, r}}$;}\\
                     c  (P_{\mathrm{load, r}} -  P_{\mathrm{load, max}}) & \text{otherwise.}
              \end{cases}
\end{displaymath}

$P_{\mathrm{load, max}}$ and $V_{\mathrm{load, mpp}}$ are obtained numerically from the following equation by iterating over the load current $i_{\mathrm{load, d}}$ (power is voltage times current):
\begin{displaymath}
V_{\mathrm{st}}  =  N_{\mathrm{s}} \left\{  E_{\mathrm{Nernst}} -   A \ln\left( \frac{\displaystyle i_{\mathrm{load, d}} / N_{\mathrm{p}} + i_{\mathrm{n, d}}}{\displaystyle i_{\mathrm{0,d}}}\right) + \right.           \left. B \ln\left(1 -  \frac{\displaystyle i_{\mathrm{load, d}} / N_{\mathrm{p}} + i_{\mathrm{n, d}}}{\displaystyle i_{\mathrm{limit, d}}}\right)  -          (i_{\mathrm{load, d}} / N_{\mathrm{p}} + i_{\mathrm{n, d}}) r_{\mathrm{a}} \right\}, 
\label{eqn.polarization}
\end{displaymath}
where $V_{\mathrm{st}}$ is the stack voltage, $E_{\mathrm{Nernst}}$ is the Nernst e.m.f.,  $A$ and $B$ are constants known from electrochemistry, $r_{\mathrm{a}}$ is the area-specific resistance, and the $i$'s represent different types of current densities (the subscript d is used to indicate density) in the cell \cite{larminie2003fuel, chakraborty2019new}.   The numerical values of the parameters are provided in Table~\ref{tab:PEMFCconstants}.

\begin{table}
\caption{Bounds of the design variables \cite{mohamed2004proton}.}
\centering
\begin{tabular}{lll}
\toprule
\textbf{Variable}   & \textbf{Lower bound}   &  \textbf{Upper bound} \\
\midrule
$N_{\mathrm{s}}$    &    1  &  50    \\
$N_{\mathrm{p}}$  &    1  &  50    \\
$A_{\mathrm{cell}}$ (cm$^2$)  & 10 & 400 \\
\bottomrule
\end{tabular}
\label{tab:bounds}
\end{table}

\begin{table}
\caption{PEMFC parameters and coefficients}
\label{tab:PEMFCconstants}
\centering
\begin{tabular}{@{\extracolsep{5pt}}ll@{}} 
\toprule                                               
\textbf{Parameter} &  \textbf{Value}  \\
\midrule
$V_{\mathrm{load, r}}$  & 12 V \\
$P_{\mathrm{load, r}}$ & 200 W \\
$K_{\mathrm{n}}$ &  0.5 \\
$K_{\mathrm{diff}}$ &  10 \\
$K_{\mathrm{a}}$  &   0.001 \\
$c$ & 200 \\
$r_{\mathrm{a}}$  &  98.0$\times$10$^{-6}$ $K\Omega$ cm$^{2}$ \\
${\displaystyle i_{\mathrm{limit, d}}}$ &  129 mA/cm$^2$ \\
${\displaystyle i_{\mathrm{0, d}}}$ &  0.21 mA/cm$^2$ \\
${\displaystyle i_{\mathrm{n, d}}}$ &  1.26 mA/cm$^2$ \\
$A$ &  0.05 V\\
$B$ &  0.08 V \\
$E_{\mathrm{Nernst}}$ &  1.04 V\\
\bottomrule
\end{tabular}
\end{table}

Tables~\ref{tab:fuel-sjaya} and \ref{tab:fuel-jaya} present results of the two algorithms on the fuel cell problem; 30 independent runs are executed for each of 13 PopSize-Gens combinations for either algorithm. 
For this problem, the success of a run is defined as the production of at least one solution with a fitness of 13.62 or lower~\cite{chakraborty2019proton}. 
For 12 of the 13 cases in Table~\ref{tab:fuel-sjaya}, the mean of the best-of-run costs is better for SJaya than for Jaya. And, on the mean FirstHitEvals metric, SJaya outperforms Jaya 10 out of the 13 times. Again, SJaya beats Jaya 9-3-1 on the success count metric. 
Results of Smith-Satterthwaite tests (Table~\ref{tab:fuel-tscores}) show that for the best-of-run cost metric, the $t$-statistic is positive in all cases but one, but the one-tailed $p$-values are not less than 0.05. Thus we do not have a strong reason at the 5\% significance level to reject the null hypothesis that the two means of the best-of-run costs are equal. For the best-of-run metric, the single negative $t$-score in Table~\ref{tab:fuel-tscores} corresponds to a $p$-value that is close to 0.5, indicating no reason to consider Jaya to be significantly better than SJaya on that case. The FirstHitEvals metric shows SJaya to be significantly better (at the 5\% level) in two of the 12 cases, the other cases being ties at that level of significance. 

Table~\ref{tab:fuel-wilcoxon} shows results of Wilcoxon signed-rank tests for the PEMFC problem.  
For each of the two metrics, the $W$-statistic is less than the critical $W$. Moreover, the one-tailed $p$-value computed from the $z$-score is less than 0.05 for both the metrics, thereby establishing a statistically significant (at the 5\% level) superiority of SJaya over Jaya on the fuel cell problem.

\begin{table*}
\caption{Results of SJaya on the fuel cell problem (each row corresponds to 30 independent runs). Most numbers are shown with rounding at the fourth place after the decimal.}
\label{tab:fuel-sjaya}
\centering
\begin{tabular}{@{\extracolsep{5pt}}ccccccccc@{}} 
\toprule
  \multirow{2}{*}{\textbf{PopSize}} & \multirow{2}{*}{\textbf{Gens}} 	& \multicolumn{3}{c} {\bf Best-of-run Fitness} & \multicolumn{4}{c} {\bf FirstHitEvals}\\
\cmidrule{3-5}  \cmidrule{6-9}                                               
                                                  &                        &  \textbf{Best} & \textbf{Mean} & \textbf{Std Dev} &  \textbf{Success} &\textbf{Best} & \textbf{Mean} & \textbf{Std Dev} \\
\midrule
20 & 10 & 13.6162  & 13.6885  & 0.0759  &    3 & 127 & 172.0 & 31.9479  \\
15 & 20 & 13.6161  & 13.6255  & 0.0190  & 21 & 128 & 254.8095 & 47.8187 \\
20 & 20 & 13.6159  & 13.6376  & 0.0523  &  21 & 127 & 310.0 & 71.8338  \\
20 & 25 & 13.6159  & 13.6302  & 0.0484  & 25 & 127 & 335.8 & 89.0222 \\
25 & 40 & 13.6157  & 13.6164  & 0.0023  & 29 & 89 & 510.6897  & 166.4118  \\
40 & 25 & 13.6158  & 13.6184  & 0.0044  &  25 & 291 & 654.24 & 213.3435  \\
20 & 100 & 13.6157  & 13.6158  & 8.7813e-05 &  30 & 127 & 436.1333 & 304.5035 \\
100 & 20 & 13.6159  & 13.6195  & 0.0029  & 20 & 463 & 1491.5 & 437.1448 \\
30 & 100 & 13.6157  & 13.6158  & 0.0002  &  30 & 370 & 585.4667 & 230.4605  \\
100 & 30 & 13.6158  & 13.6179  & 0.0022  &  25 & 463 & 1675.08 & 550.3110  \\
40 & 100 & 13.6157  & 13.6160  & 0.0006  & 30 & 291 & 778.9333  & 385.4906  \\
100 & 40 & 13.6157  & 13.6174  & 0.0022  & 26 & 463 & 1737.1154 & 622.4179  \\ 
100 & 100 & 13.6157  & 13.6162  & 0.0010  &  29 & 463 & 2155.3103  & 1395.2800  \\
\bottomrule
\end{tabular}
\end{table*}

\begin{table*}
\caption{Results of Jaya on the fuel cell problem (each row corresponds to 30 independent runs). Most numbers are shown with rounding at the fourth place after the decimal.}
\label{tab:fuel-jaya}
\centering
\begin{tabular}{@{\extracolsep{5pt}}ccccccccc@{}} 
\toprule
  \multirow{2}{*}{\textbf{PopSize}} & \multirow{2}{*}{\textbf{Gens}} 	& \multicolumn{3}{c} {\bf Best-of-run Fitness} & \multicolumn{4}{c} {\bf FirstHitEvals}\\
\cmidrule{3-5}  \cmidrule{6-9}                                               
                                                  &                        &  \textbf{Best} & \textbf{Mean} & \textbf{Std Dev} &  \textbf{Success} &\textbf{Best} & \textbf{Mean} & \textbf{Std Dev} \\
\midrule

20 & 10 & 13.6213  & 13.7026  & 0.0713  & 0 & --- & --- \\
15 & 20 & 13.6160  & 13.6374  & 0.0342  & 13 & 124 & 241.8462 & 45.9378 \\
20 & 20 & 13.6163  & 13.6367 & 0.0483 & 20 & 298 & 363.65 & 31.7636  \\
20 & 25 & 13.6160 & 13.6312 & 0.0463  & 25 & 298 & 382.36 & 48.3867  \\
25 & 40 & 13.6158  & 13.6298 & 0.0520  & 28 & 144 & 540.6071 & 144.4191  \\
40 & 25 & 13.6158  & 13.6229  & 0.0236  & 26 & 250 & 739.5 & 170.0993  \\
20 & 100 & 13.6157 & 13.6182  & 0.0126 & 29 & 298 & 454.6897  & 236.2226 \\
100 & 20 & 13.6160  & 13.7947  & 0.9338  & 14 & 907 & 1595.2857  & 360.2738 \\
30 & 100 & 13.6157  & 15.1444  & 8.2308  &  29 & 368 & 740.6207 & 546.2285  \\
100 & 30 & 13.6159  & 13.7910  & 0.9344  &  25 & 907 & 1922.44 & 492.2595  \\
40 & 100 & 13.6157 & 13.6202  & 0.0237  &  29 & 250 & 787.5517  & 222.2544 \\
100 & 40 & 13.6157  & 13.7907  & 0.9345  & 26 & 907 & 1972.7308  & 544.2687 \\
100 & 100 & 13.6157  & 13.7900  & 0.9346 &  27 & 907 & 2118.4074  & 914.8884 \\
\bottomrule
\end{tabular}
\end{table*}


\begin{table*}
\caption{Smith-Satterthwaite tests: Jaya vs. Sjaya on the fuel cell problem}
\label{tab:fuel-tscores}
\centering
\begin{tabular}{@{\extracolsep{5pt}}cccccc@{}} 
\toprule
\multirow{2}{*}{\textbf{PopSize}} & \multirow{2}{*}{\textbf{Gens}} 	& \multicolumn{2}{c} {\bf Best-of-run Fitness} & \multicolumn{2}{c} {\bf FirstHitEvals}\\
\cmidrule{3-4}  \cmidrule{5-6}                                               
                                             &                        &  \textbf{$t$-statistic} &  \textbf{$p$-value} & \textbf{$t$-statistic} & \textbf{$p$-value} \\
\midrule
20 & 10 & 0.7429 & 0.2303 & --- & --- \\
15 & 20 & 1.6673  & 0.0512  &  -0.7872  & 0.2191  \\
20 & 20 &  -0.0627  & 0.4751  &  3.1175 &  0.0021  \\
20 & 25 & 0.0865  & 0.4657  &  2.2976  & 0.0137 \\
25 & 40 & 1.4068  & 0.0850  &  0.7256  & 0.2356  \\
40 & 25 & 1.0202 & 0.1578  &  1.5742 & 0.0612  \\
20 & 100 & 1.0461  &  0.1521   &  0.2620 & 0.3971  \\
100 & 20 & 1.0279  & 0.1562  &  0.7564  &  0.2276  \\
30 & 100 &  1.0172  & 0.1587 & 1.4129 & 0.0830 \\
100 & 30 & 1.0147  & 0.1593  & 1.6751 &  0.0503 \\
40 & 100 & 0.9838  &  0.1667  & 0.1056 & 0.4582 \\
100 & 40 & 1.0158 & 0.1591 & 1.4530 & 0.0763  \\
100 & 100 & 1.0190 &  0.1583 &  -0.1178 &  0.4534 \\
\bottomrule
\end{tabular}
\end{table*}

\begin{sidewaystable}
\caption{Wilcoxon signed rank tests: Jaya vs Sjaya on the fuel cell problem.}
\centering
\begin{tabular}{p{4.5cm} p{1cm} p{1cm} p{1cm} p{1cm} p{1cm} p{1cm} p{1cm} p{1cm} p{1.5cm} p{1.5cm} p{1cm}}
\toprule
\textbf{Metric} & \textbf{\#zero diff.} & \textbf{$n$} & \textbf{$W+$} &  \textbf{$W-$} &  \textbf{$W$} & \textbf{$\alpha$} & \textbf{Critical $W$} & \textbf{Mean of $W$} & \textbf{Std. Dev. of $W$} & \textbf{$z$-statistic} & \textbf{left tail $p$} \\ 
\midrule
Mean of Best-of-Run Fitnesses  & 0 & 13 & 90 & 1 & 1 & 0.05 & 21 & 45.5 & 14.3091 & $-$3.1099 & 0.0009 \\
Mean of FirstHitEvals               & 0 & 12 & 71 & 7 & 7 & 0.05 & 17 & 39    & 12.7475 & $-$2.5103  & 0.0060 \\
\bottomrule
\end{tabular}
\label{tab:fuel-wilcoxon}
\end{sidewaystable}

\section{Conclusions}
\label{sec:conclu}
This paper presented an improvement to the Jaya algorithm by introducing new update policies in the search process. The usefulness of the present approach is that, unlike most other improvements to Jaya reported in the literature, our strategy does not require the introduction of any additional parameter. It retains both the features that the original Jaya is famous for, namely ``parameterlessness" and simplicity, while providing performance that is statistically significantly better (in terms of the solution quality) and/or faster (in terms of the speed of finding a near-optimal solution) than that produced by Jaya.

\bibliographystyle{ieeetr}
\bibliography{udaybib}

\end{document}